%% file: main.tex
\documentclass[journal]{vgtc}                     

\onlineid{0}

\vgtccategory{Research}

\vgtcpapertype{system}

\title{HarassGuard: Detecting Harassment Behaviors in Social Virtual Reality with Vision-Language Models}

\author{%
  \authororcid{Junhee Lee}{0009-0009-4967-3621},
  \authororcid{Minseok Kim}{0009-0004-3991-6486},
  \authororcid{Hwanjo Heo}{0000-0002-8105-4224},
  \authororcid{Seungwon Woo}{0009-0001-8874-6936},
  \authororcid{Jinwoo Kim}{0000-0003-1303-8668}*
}

\authorfooter{
  \item
  	Junhee Lee is with Kwangwoon University. E-mail: hunroung0202@gmail.com
  \item
  	Minseok Kim is with Kwangwoon University.
  	E-mail: rhdtls1000@kw.ac.kr

  \item 
    Hwanjo Heo is with ETRI.
    E-mail: hwanjo@etri.re.kr

  \item
    Seungwon Woo is with ETRI.
    E-mail: seungww@etri.re.kr

  \item
    Jinwoo Kim is with Chungbuk National University.
    E-mail: jinwookim@cbnu.ac.kr (*Corresponding author. This work was primarily performed while the author was with Kwangwoon University.)
}

\input{sections/abstract}


\teaser{
  \centering
  \includegraphics[width=0.8\linewidth]{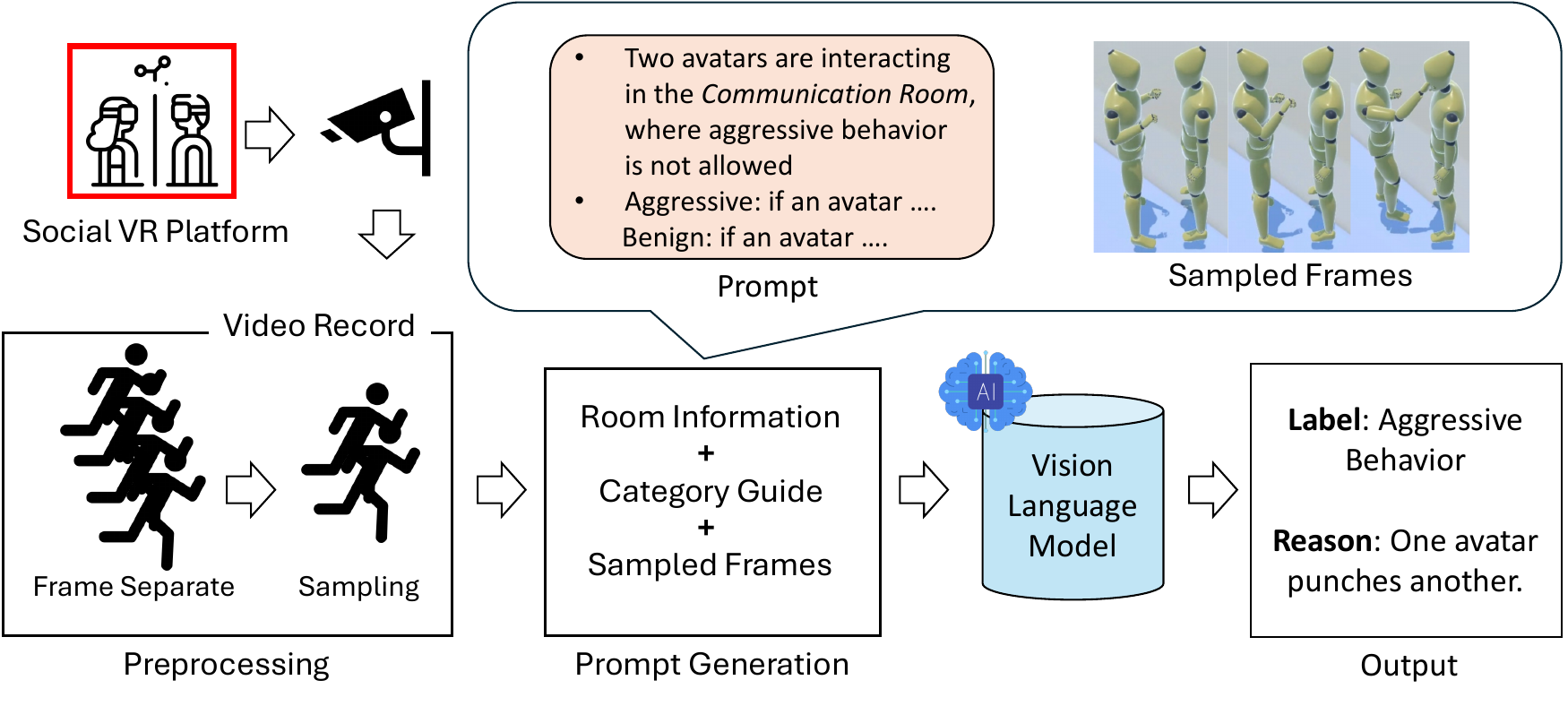}
  \caption{The overall architecture of \ourtool{}, which detects harassment behavior in social VR by leveraging a vision-language model (VLM).}
  \label{fig:overview}
}

\renewcommand{\manuscriptnotetxt}{This article has been accepted for publication in the IEEE Transactions on Visualization and Computer Graphics (TVCG) Special Issue on the 2026 IEEE Conference on Virtual Reality and 3D User Interfaces (VR), and was presented at IEEE VR 2026.}

\usepackage{tabu}                      
\usepackage{booktabs}                  
\usepackage{lipsum}                    
\usepackage{mwe}                       

\usepackage{mathptmx}                  

\newcommand{\ourtool}{\texttt{HarassGuard}}

\newif\ifhighlight
 \highlightfalse 

\ifhighlight
    
    \newcommand{\revc}[1]{\textcolor{blue}{\st{#1}}} 
\else
    \newcommand{\revc}[1]{}   
\fi

\usepackage{kotex}
\usepackage{listings}
\usepackage{multicol}
\usepackage{multirow}
\usepackage{tcolorbox}
\usepackage{xurl}
\usepackage{soul}

\begin{document}


\firstsection{Introduction}

\maketitle

\input{sections/intro}

\input{sections/background}
\input{sections/design}

\input{sections/eval}

\input{sections/relatedwork}
\input{sections/conclusion}


\bibliographystyle{abbrv-doi-hyperref}

\bibliography{references}








\end{document}

%% file: sections/abstract.tex
\abstract{
Social Virtual Reality (VR) platforms provide immersive social experiences but also expose users to serious risks of online harassment. Existing safety measures are largely reactive, while proactive solutions that detect harassment behavior during an incident often depend on sensitive biometric data, raising privacy concerns. In this paper, we present \ourtool{}, a vision-language model (VLM) based system that detects physical harassment in social VR using only visual input. We construct an IRB-approved harassment vision dataset, apply prompt engineering, and fine-tune VLMs to detect harassment behavior by considering contextual information in social VR. Experimental results demonstrate that \ourtool{} achieves competitive performance compared to state-of-the-art baselines (i.e., LSTM/CNN, Transformer), reaching an accuracy of up to 88.09\% in binary classification and 68.85\% in multi-class classification. Notably, \ourtool{} matches these baselines while using significantly fewer fine-tuning samples (200 vs. 1,115), offering unique advantages in contextual reasoning and privacy-preserving detection.
}

%% file: sections/intro.tex

Over the past decade, social virtual reality (VR) platforms have gained significant popularity due to their immersive virtual experiences. Users can participate in a wide range of social events, games, and 3D virtual worlds, interacting through personalized avatars, which makes them more engaged than with traditional social networking services. VRChat, a representative social VR platform, has recently recorded about 60,000 concurrent users~\cite{vrchat_news}.

Nevertheless, the success of social VR platforms also reveals a darker side: \emph{online harassment}, which has become a serious issue both within VR communities and in the real world. While online harassment has long been a concern on traditional social networking services, the immersive experiences and sense of embodiment provided by social VR platforms make incidents such as stalking, physical intrusion, and sexual assault potentially even more harmful to users’ mental health~\cite{freeman2022disturbing,abhinaya2024enabling}. A recent report~\cite{bbc_sexual} noted that police investigated a virtual sexual assault incident, underscoring that this is not merely a virtual issue but one with real-world consequences.

To mitigate online harassment on social VR platforms, various safety measures have been proposed, such as personal boundaries (or “bubbles”) and tools for blocking or reporting harassers~\cite{freeman2022disturbing}. However, this \emph{reactive} approach depends on the victim’s response after the harassment has already occurred, which is problematic since harassment often happens suddenly, leaving victims little time to respond. Alternatively, some studies have explored detecting harassing behaviors using deep learning models that monitor users’ actions by collecting biometric information (e.g., controller inputs and avatar positions)~\cite{wang2024hardenvr}. While this \emph{proactive} approach can address the limitations of reactive measures, it also raises significant concerns about user privacy, as such biometric data could be exploited in privacy leakage attacks~\cite{meng2023anonymization,nair2023unique,pfeuffer2019behavioural}.

In this paper, we aim to address this limitation by proposing an approach for detecting harassment behavior\footnote{For simplicity, we focus solely on \emph{physical harassment}, as it represents a particularly serious threat in social VR due to the immersive nature of these environments.} in social VR, based exclusively on \emph{visual input} rather than on the collection of biometric information. A naïve approach might attempt to leverage well-established vision models specialized for human action recognition (HAR), training them to identify harassment behaviors captured from social VR platforms. While this goal may seem straightforward, it poses several technical challenges. 

First, harassment behavior datasets specific to social VR platforms are limited, making it difficult to effectively fine-tune vision models. Although existing real-world HAR datasets~\cite{mukherjee2017fight,soomro2012ucf101,soliman2019violence} could be leveraged, social VR feature a wide variety of avatars that differ significantly from real-human appearances, and thus directly applying such datasets can significantly degrade detection accuracy. Second, harassment behaviors in social VR are often hard to distinguish from benign actions. For example, in a communication room~\cite{unity_multiplayertemplate}, a user's light touch for a greeting can be visually similar to an aggressive strike, which traditional vision models might erroneously flag as harassment. Thus, it is crucial to incorporate \emph{contextual information}, such as the theme of the virtual world, user relationships (e.g., whether they are friends), and the duration of the behavior, to improve detection reliability. However, providing vision models with such multimodal inputs is non-trivial, as it requires integrating diverse data sources and designing architectures capable of effectively utilizing them.

We address these limitations through the following approaches. First, we construct a harassment vision dataset specific to social VR from our IRB-approved user study. In this study, participants are instructed to simulate not only harassment behaviors based on our proposed taxonomy (e.g., aggressive behavior, personal space violations, disruptive behavior) but also benign behaviors in a custom virtual world, using a single avatar type for the proof-of-concept study. Second, we feed this dataset to vision-language models (VLMs), which can process multimodal inputs (e.g., text, images) to accurately detect harassment behavior within social VR platforms. To further improve detection performance, we employ two strategies: (i) we apply prompt engineering to guide the VLM in reasoning about the context of input VR video frames, and (ii) we fine-tune the VLM with our constructed dataset to help it capture the inherently vague nature of harassment behaviors in social VR environments.

We evaluate \ourtool{} with GPT-4o as a backend VLM on the constructed dataset and demonstrate its effectiveness in performing vision-based harassment detection. By comparing against baselines, such as LSTM/CNN- and Transformer-based models, we show that our prompt-engineered and fine-tuned VLM can achieve performance comparable to these baselines, reaching an accuracy of up to 88.09\% in binary classification and 68.85\% in multi-class classification. Notably, \ourtool{} achieves these results even when trained on a relatively small dataset (200 samples, compared to 1,115 for the baselines), highlighting its potential for data-efficient adaptation to diverse social VR scenarios.

\noindent\textbf{Contributions.} Our contributions are summarized as follows:

\begin{itemize}
    \item We constructed a harassment vision dataset through an IRB-approved user study on our custom social VR platform. To promote open science, the dataset will be made available upon request.  
    \item We designed and implemented \ourtool{}, a prototype VLM-based harassment detection system, enhanced through prompt engineering and fine-tuning.  
    \item We evaluated \ourtool{} against conventional vision models and demonstrated that it can accurately and efficiently detect harassment behaviors in social VR.  
\end{itemize}

%% file: sections/background.tex
\section{Background and Related Work} 

In this section, we provide the necessary background and review related work relevant to our study.

\subsection{Harassment in Social VR} 

Online harassment has long been recognized as a serious threat, not only in conventional social networking services (e.g., Facebook) but also in Internet forums. Harassers often engage in hate speech, as well as racist and sexist remarks, which cause significant psychological harm to victims~\cite{porta2024sexual,jadamba2025there}. More recently, harassment behaviors have emerged as a critical issue in social virtual reality (VR) platforms especially for young users~\cite{hinduja2024metaverse}. Unlike traditional platforms, VR offers immersive and embodied experiences through avatar self-similarity~\cite{tschanter2025towards}, making harassment feel more realistic and distressing.
However, preventing harassment in VR remains challenging because perceptions of harassment are often subjective, making it difficult to apply platform governance consistently~\cite{blackwell2019harassment}.
This threat is not merely hypothetical; for example, police have investigated cases of sexual harassment targeting girls’ avatars~\cite{bbc_sexual}, and Interpol is actively developing training programs for policing metaverse environments~\cite{interpol}. 

\subsection{Existing Solutions}
\label{subsec:existing_solutions}

Recently, numerous studies on harassment in social VR have been presented, and various solutions have been discussed in both the HCI and security communities. Deldari et al.~\cite{deldari2023investigation} conducted a user study with victims on actual social VR platforms, reporting that harassment—including sexual harassment—was frequently observed in private rooms of platforms primarily used by teenagers. The study emphasized the need for moderators to prevent such incidents; however, it found that platforms generally provide inadequate support in this regard. Abhinaya et al.~\cite{abhinaya2024enabling} interviewed developers of social VR platforms and revealed that, while platforms are aware of harassment issues, they struggle to address them due to limited technical resources and manpower, and these issues are often not treated as a primary priority. In Meta Horizon Worlds, safety specialists have been deployed to monitor and record in-world behaviors~\cite{horizon_moderator}, and a personal boundary (``bubble'') feature has been introduced to block unwanted approaches from other users~\cite{horizon_bubble}. Nevertheless, such measures have not yet been adopted across other social VR platforms.

To address this issue, several technical solutions have recently been proposed. Wang et al.~\cite{wang2024hardenvr} introduced a deep learning-based method called HardenVR for detecting harassment behaviors in social VR platforms. Specifically, they collected biometric information from users’ avatars, including avatar position, hand position, and controller button inputs, and used these features to train an LSTM (Long Short-Term Memory) model. Their system achieved over 98\% accuracy under their experimental setting in detecting harassment motions such as pinching, punching, and slapping. However, the collection of such biometric data raises privacy concerns, as it relies on high-precision motion or physiological data that can function as permanent digital identifiers of a user's physical body~\cite{meng2023anonymization,pfeuffer2019behavioural,nair2023unique}. Additionally, most social VR platforms still rely primarily on user reports and manual recording of harassment incidents, where users have expressed dissatisfaction with these measures, noting that reports and evidence collection often fail to result in adequate action~\cite{weerasinghe2025beyond}. Finally, their evaluation results reveal a critical limitation: detection accuracy is markedly higher when controller inputs are included, but drops significantly when relying solely on avatar and hand positional data. This indicates that classification performance relies heavily on controller-based biometric features, making the system less effective in scenarios where only positional information is available. 

To overcome these shortcomings, we conducted a user study to curate a new dataset specifically for harassment detection. Unlike prior datasets that rely on coordinate- or controller-based modalities, our dataset captures vision-based behavioral cues, enabling the identification of harassment through observable actions. This approach offers a more privacy-conscious alternative by avoiding the collection of raw biometric markers and also allows harassment evidence to be communicated more transparently to both platform providers and end users, as visual representations combined with vision-language model (VLM) reasoning provide an intuitive explanation of abusive behaviors. Importantly, this direction aligns with recent survey findings~\cite{weerasinghe2025beyond} showing that users want to better understand how platforms handle harassment cases, including why perpetrators were processed in particular ways and on what grounds such decisions were made. Additionally, our approach can also be applied to other avatar types by adjusting the semantic information of different avatar characteristics when fed to VLM, because VLM can adaptively understand subtle differences between avatar appearance through fine-tuning with a relatively small dataset, while previous raw data-based solutions do not.

\subsection{Threat Model and Assumptions}

We aim to detect harassment behaviors in social VR platforms. We assume the presence of a \textit{harasser} who directs harassment toward a \textit{victim}. Detecting such incidents requires visual evidence, which can be obtained in two ways: (i) the social VR platform provides cameras attached to all users, or (ii) a \textit{bystander} records the interaction. In this paper, we adopt the first assumption, though our approach is equally applicable to the latter.

%% file: sections/design.tex
\section{User Study}
\label{sec:user_study}

In this section, we present the user study and the resulting dataset aimed at capturing vision-based recordings of harassment behaviors in social VR environments. Existing studies do not utilize vision-based data; even when such modalities are employed, harassment specific datasets are rarely available. Most prior work focuses on detecting specific actions such as throwing objects or on tasks related to de-anonymization, using head, controller, or gaze data~\cite{kupin2018task,miller2020personal,meng2023anonymization,moore2021personal,pfeuffer2019behavioural,tricomi2209you,miller2023large}. While HardenVR~\cite{wang2024hardenvr} provides a harassment dataset, it relies on positional (coordinate) information and thus cannot fully support vision-based analysis. To address this gap, we recruited 14 participants and conducted a user study to create a new vision-based dataset, accompanied by a survey to collect participants’ perspectives on harassment behaviors. The study was reviewed and approved by the Institutional Review Board (IRB) of Kwangwoon University (No. 7001546-20250630-HR(SB)-006-06), and informed consent was obtained from all participants prior to the experiment.

\subsection{Social VR Platform Setup}
To facilitate the implementation of our social VR platform, we utilized the Unity VR Multiplayer Template~\cite{unity_multiplayertemplate}. Since this template targets social VR platforms such as VRChat, we used Unity version 2022.3.22f1, which corresponds to VRChat’s development environment. The default Unity template features avatars displaying only the upper body; to create an environment more similar to VRChat, we employed a free full-body avatar from the Unity Asset Store~\cite{unity_banana} and leveraged the Animation Rigging package to implement natural upper-body movements. For the lower body, we developed an inverse kinematics script to reproduce movements comparable to those in social VR platforms. To ensure natural interaction, the platform supported standard social VR features, including upper-body tracking via headset and controllers, and object interaction using the grip button (see Fig. 6 in our supplementary material).

We deployed the social VR platform on a high-performance Windows server, which participants accessed through a custom Unity application. To enable data collection, the system recorded all video streams directly on the server. Four cameras provided a third-person, omnidirectional view of participants, following them as they moved, while an additional camera captured the participant’s first-person perspective. This setup was strictly for data collection purposes and was not intended to simulate any harassment detection functionality within a real social VR platform.

\subsection{Harassment Behavior Classification}
\begin{table}[t]
\small
\centering
\caption{Categorization of harassment and benign behaviors in social VR for our user study.}
\label{tab:category}
\begin{tabular}{>{\centering}m{2.4cm} m{2cm} m{3cm}}
\toprule
\textbf{Category} & \textbf{Subcategory} & \textbf{Definition} \\
\midrule
\multirow{6}{*}{Aggressive Behavior} 
 & Punching & Attacking another avatar with a fist \\ \cmidrule{2-3}
 & Slapping & Attacking another avatar with an open hand \\ \cmidrule{2-3}
 & Hitting with object & Attacking anothe avatar using an object \\ \midrule
\multirow{3}{*}{Personal Space Violation} 
 & Looming & Bringing one's avatar face close to another user's avatar \\ \cmidrule{2-3}
 & Following / Stalking & Tracking a user repeatedly over time \\ \midrule
Disruptive Behavior & Blocking & Obstructing the movement of another avatar \\ \midrule
Benign Behavior & - & All other benign behaviors (e.g., talking, rock-paper-scissors, walking) \\ \bottomrule
\end{tabular}
\vspace{-0.1in}
\end{table}

\begin{figure*}[!t]
    \centering
    \begin{subfigure}[b]{0.25\linewidth}
        \centering
        \includegraphics[width=\linewidth]{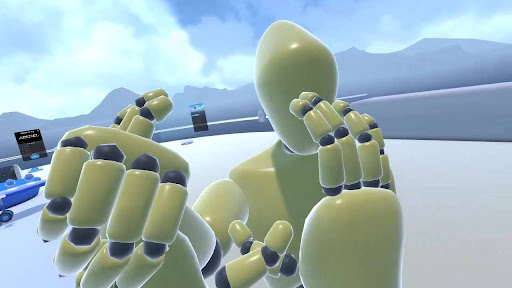}
        \caption{Example of a Punching Action}
        \label{fig:exam-punching}
    \end{subfigure}
    \begin{subfigure}[b]{0.25\linewidth}
        \centering
        \includegraphics[width=\linewidth]{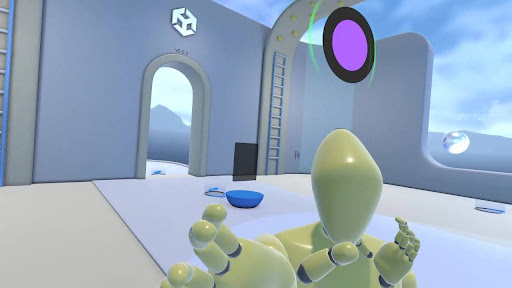}
        \caption{Example of a Slapping Action}
        \label{fig:exam-slapping}
    \end{subfigure}
    \begin{subfigure}[b]{0.25\linewidth}
        \centering
        \includegraphics[width=\linewidth]{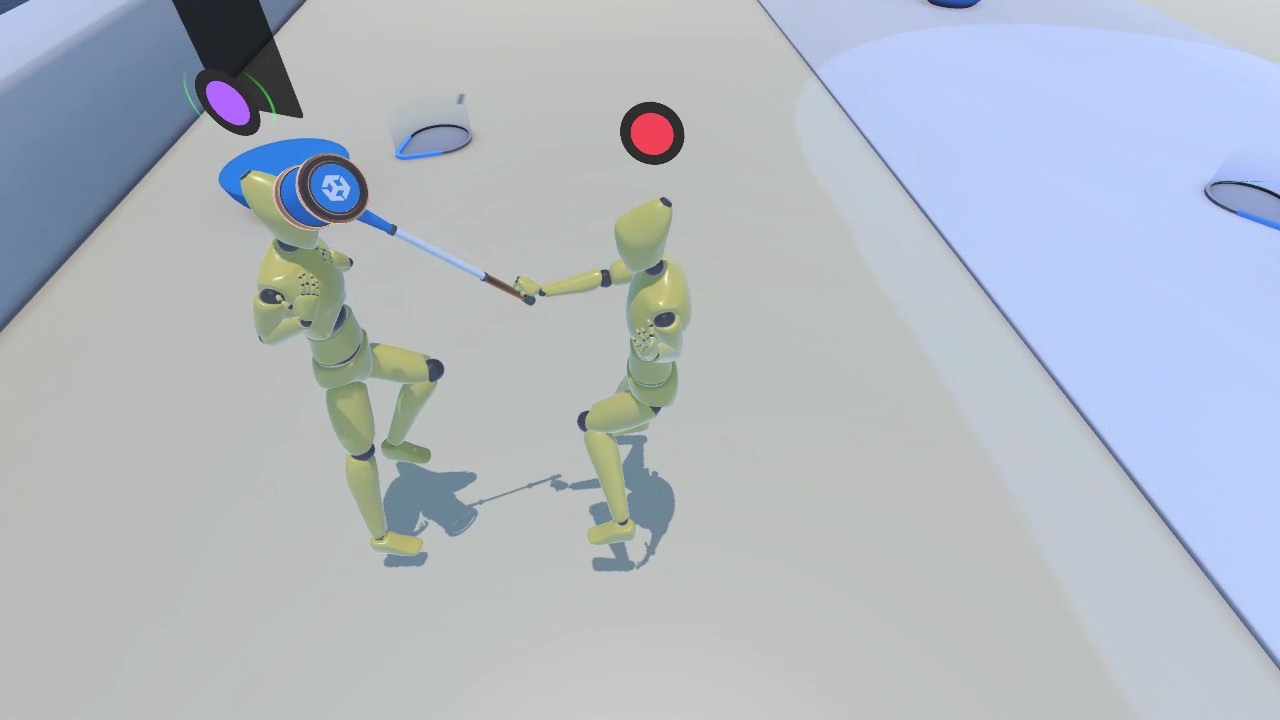}
        \caption{Example of a Hitting With Object Action}
        \label{fig:exam-objects}
    \end{subfigure}
    \begin{subfigure}[b]{0.25\linewidth}
        \centering
        \includegraphics[width=\linewidth]{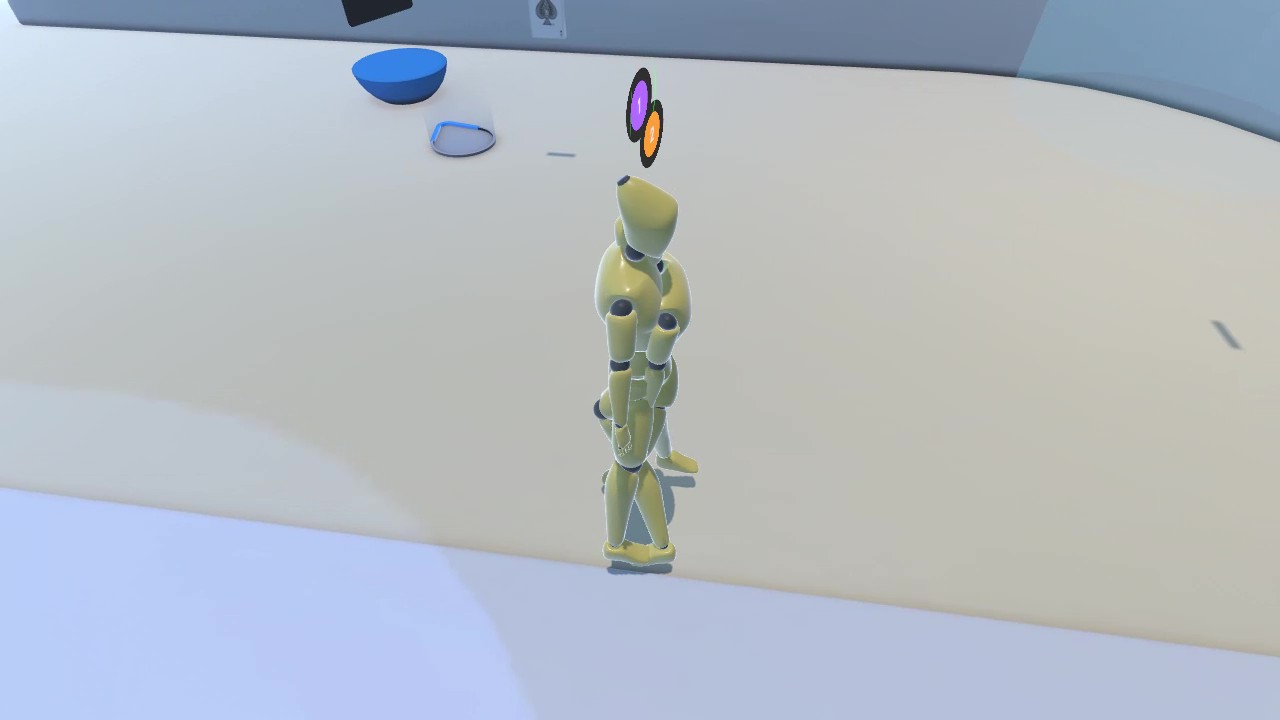}
        \caption{Example of a Looming Action}
        \label{fig:exam-looming}
    \end{subfigure}
    \begin{subfigure}[b]{0.25\linewidth}
        \centering
        \includegraphics[width=\linewidth]{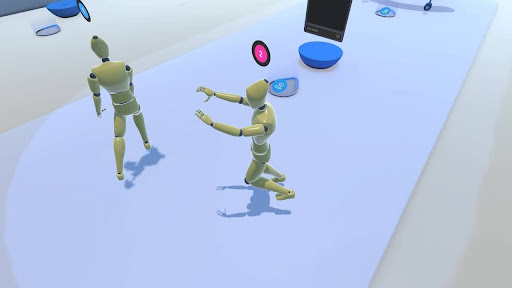}
        \caption{Example of a Following / Stalking Action}
        \label{fig:exam-following}
    \end{subfigure}
    \begin{subfigure}[b]{0.25\linewidth}
        \centering
        \includegraphics[width=\linewidth]{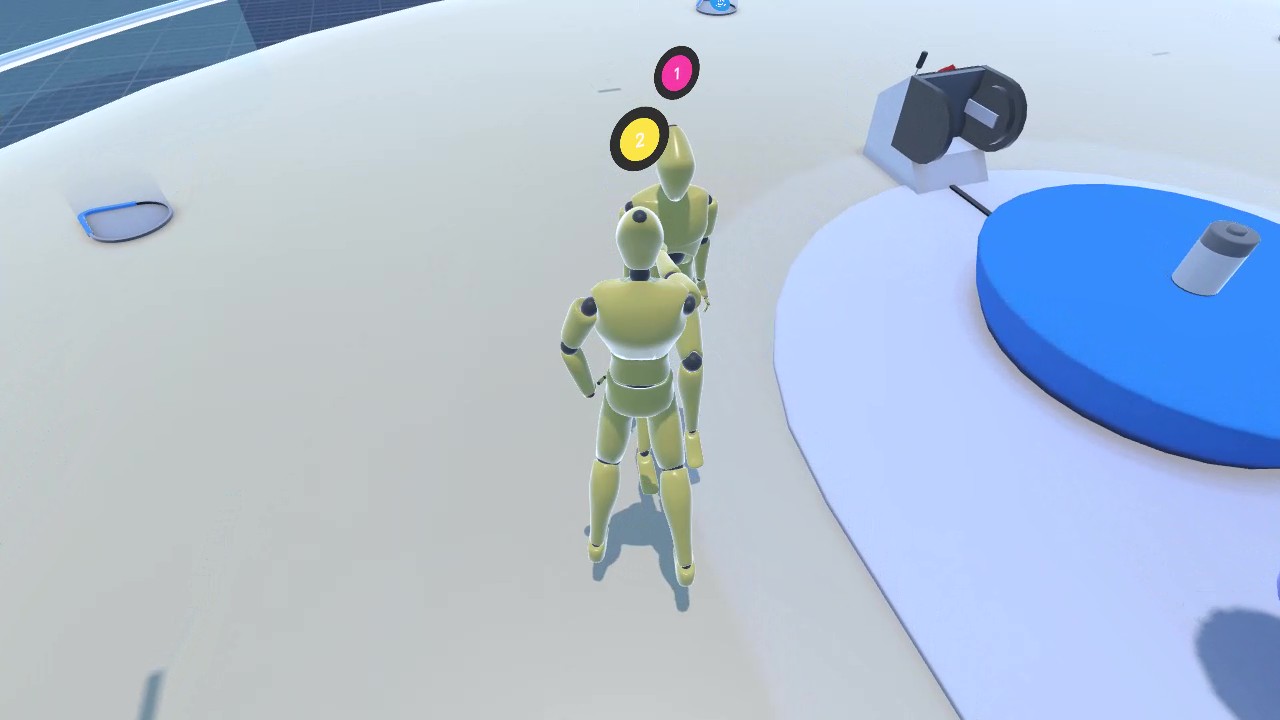}
        \caption{Example of a Blocking Action}
        \label{fig:exam-blocking}
    \end{subfigure}
    \caption{Examples of actions in Aggressive Behavior, Personal Space Violation, and Disruptive Behavior.}
    \label{fig:category}
    \vspace{-0.2in}
\end{figure*}

Table~\ref{tab:category} categorizes harassment behaviors collected in our user study. Each major category was derived by reinterpreting the results of a user study conducted by Weerasinghe et al.~\cite{weerasinghe2025beyond}, which surveyed approximately 100 participants with prior harassment experience in social VR platforms. In their study, harassment was classified into three types: verbal harassment, visual harassment, and physical harassment. In this paper, we focus specifically on harassment through physical actions. We further refined the classification by including the most unpleasant physical behaviors reported in their study (e.g., unwanted contact, violent actions) as well as other possible behaviors~\cite{blackwell2019harassment, bailenson2003interpersonal, wilcox2006personal, slater2009place, yee2007proteus}, resulting in four categories: Aggressive Behavior, Personal Space Violation, Disruptive Behavior, and Benign Behavior.

\textbf{Aggressive Behavior} refers to actions in which a user physically threatens another user’s avatar. The subcategories were defined based on clear physical actions, including punching, slapping, and hitting with objects (Fig.~\ref{fig:exam-punching}, \ref{fig:exam-slapping}, and \ref{fig:exam-objects}). Such behaviors are particularly impactful in VR due to high levels of place illusion and plausibility, which can trigger realistic psychological and behavioral responses even though the interaction is virtual~\cite{slater2009place}. Furthermore, users’ behavioral tendencies can be influenced by the Proteus effect, where their own avatar representation may amplify aggressive tendencies~\cite{yee2007proteus}.

\textbf{Personal Space Violation} describes instances in which a user approaches another user’s avatar without inflicting physical harm. Subcategories include excessive proximity (Looming, Fig.~\ref{fig:exam-looming}) and Following/Stalking (Fig.~\ref{fig:exam-following}). Prior work has demonstrated that interpersonal distance in immersive virtual environments is regulated similarly to real life, with users exhibiting discomfort or defensive responses when personal space is breached~\cite{bailenson2003interpersonal, wilcox2006personal}. This validates the importance of considering spatial interactions when classifying harassment behaviors.

\textbf{Disruptive Behavior} refers to actions that interfere with another user’s interactions or prevent them from engaging with other elements of the environment. The subcategory includes blocking (Fig.~\ref{fig:exam-blocking}), which obstructs another user’s movement. Although blocking can be implemented in various ways across social VR platforms, common approaches involve using the avatar’s physical presence or moving a physical object to impede motion. Prior studies on social VR harassment highlight that such disruptive behaviors can cause discomfort, feelings of intimidation, and frustration~\cite{blackwell2019harassment}.

\textbf{Benign Behavior} includes all behaviors not classified under the previously defined categories. To collect data for this category, participants engaged in normal gameplay and simple social interactions (e.g., rock-paper-scissors, greetings, casual conversation) within four designated rooms.

\subsection{Experiment Scenario}
Participants took part in the study in groups of 2--4 and were placed in four types of rooms that represent common environments in social VR platforms: a Communication Room, a Climbing Room, a Whack-a-Pig Room, and a Sling Shot Room. Before the experiment, participants received approximately 5 minutes of instructions on the VR headset and controllers, followed by about 15 minutes of experience across the four rooms. They were also briefed on detailed descriptions of the four types of behaviors prior to the study.

The experiment followed a within-subject design, where all participants performed both attacker and victim roles at least once across the four rooms and enacted each type of harassment behavior at least once. Participants assigned to the attacker role were allowed to freely perform harassment during the scenarios, but they were required to strictly follow the definitions provided in Table~\ref{tab:category}. Also, to distinguish between specific aggressive actions, participants were given explicit control instructions where Punching required holding both the Grip and Trigger buttons to simulate a fist, while Slapping was performed with the controllers in an idle state to represent an open hand (see Fig. 6 in our supplementary material). Note that only one type of harassment could be performed at a time. Participants were not fixed to a single group; instead, they joined sessions in varying group sizes (from two to four) and were mixed across different runs of the experiment. Including setup and experience, each group spent approximately one hour in total. After completing all scenarios, participants filled out a post-experiment survey and received a reward of approximately \$10.

\noindent\textbf{Scenario 1: Communication Room.}
The Communication Room emulates a typical social VR space where one of the most common activities is interacting and conversing with others in a well-designed environment. We placed small playing cards throughout the room, encouraging participants to move around and locate them. In this scenario, two to four participants enacted harassment behaviors according to predefined categories. While victims engaged in observation, conversation, or rock-paper-scissors, attackers performed \textit{Aggressive Behaviors} (e.g., punching, slapping) or \textit{Personal Space Violations} (e.g., stalking, looming). Uniquely, this scenario included \textit{Disruptive Behavior} (e.g., blocking), where attackers used objects to physically block victims' movements. \textit{Benign Behavior} involved standard social interactions without interference.

\noindent\textbf{Scenario 2: Whack-a-Pig Room.}
The Whack-a-Pig Room simulates a whack-a-mole–style game, where participants use a hammer to hit pigs that appear randomly. This setting was designed to collect data on whether object-hitting actions were directed at another avatar or were part of normal gameplay. Two to four participants enacted harassment behaviors according to predefined categories. To distinguish game actions from harassment, victims focused on gameplay, while attackers directed \textit{Aggressive Behaviors} (e.g., hitting with the hammer, punching, slapping) toward them. \textit{Personal Space Violations} involved visual obstruction (e.g., looming). \textit{Benign Behavior} consisted of normal gameplay.

\noindent\textbf{Scenario 3: Sling Shot Room.}
The Sling Shot Room was based on a Unity template where participants pull a slingshot to hit a target. Like the Whack-a-Pig and Climbing Rooms, this represents competitive interactions commonly found in social VR platforms. This room was used to collect data for detecting actions that interfere with another user’s immersion, beyond the intended gameplay. Two to four participants performed harassment behaviors toward one another according to the defined categories. Victims attempted normal gameplay while attackers interfered via \textit{Aggressive Behaviors} (e.g., punching, slapping) or \textit{Personal Space Violations} (e.g., looming). \textit{Benign Behavior} involved uninterrupted gameplay.

\noindent\textbf{Scenario 4: Climbing Room.}
The Climbing Room is based on a Unity template that allows avatars to interact and climb a rock wall. It was modified to enable smooth interaction with the avatars used in this study and represents a game-style room commonly found on social VR platforms. This room was used solely to collect benign behavior data; no harassment behaviors were performed, and participants freely explored the room.

\subsection{Participants}

\begin{figure}[!t]
    \centering
    \subfloat[Average VR usage time]{
        \centering
        \includegraphics[width=0.47\linewidth]{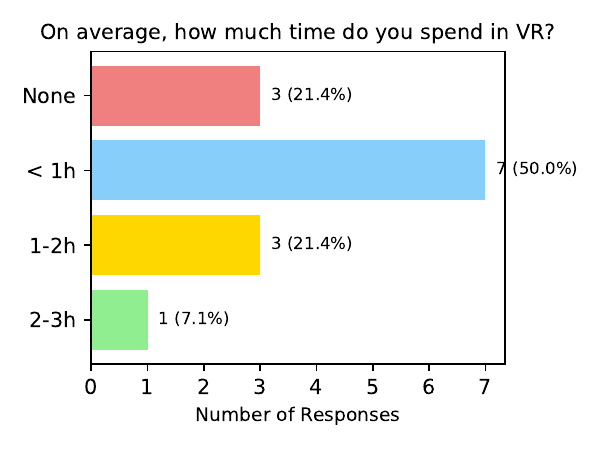}
        \label{fig:0-1}
    }
    \subfloat[Users’ responses on platform harassment actions]{
        \centering
        \includegraphics[width=0.47\linewidth]{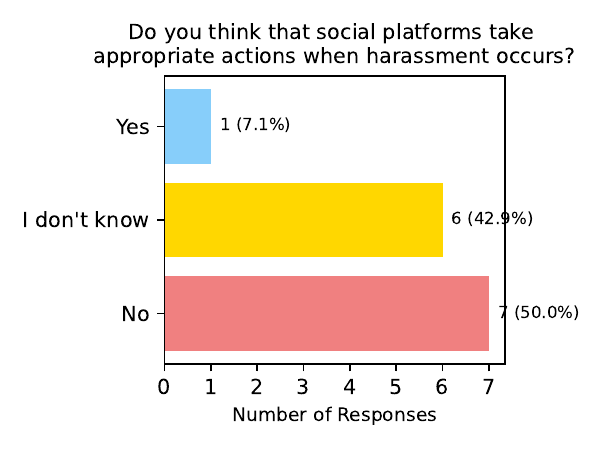}
        \label{fig:1-2}
    }
    \caption{Participants’ VR usage time and awareness of harassment responses in social platforms.}
    \label{fig:participant}
    \vspace{-0.1in}
    
\end{figure}

\begin{figure*}[!t]
    \centering
    \includegraphics[width=.9\linewidth]{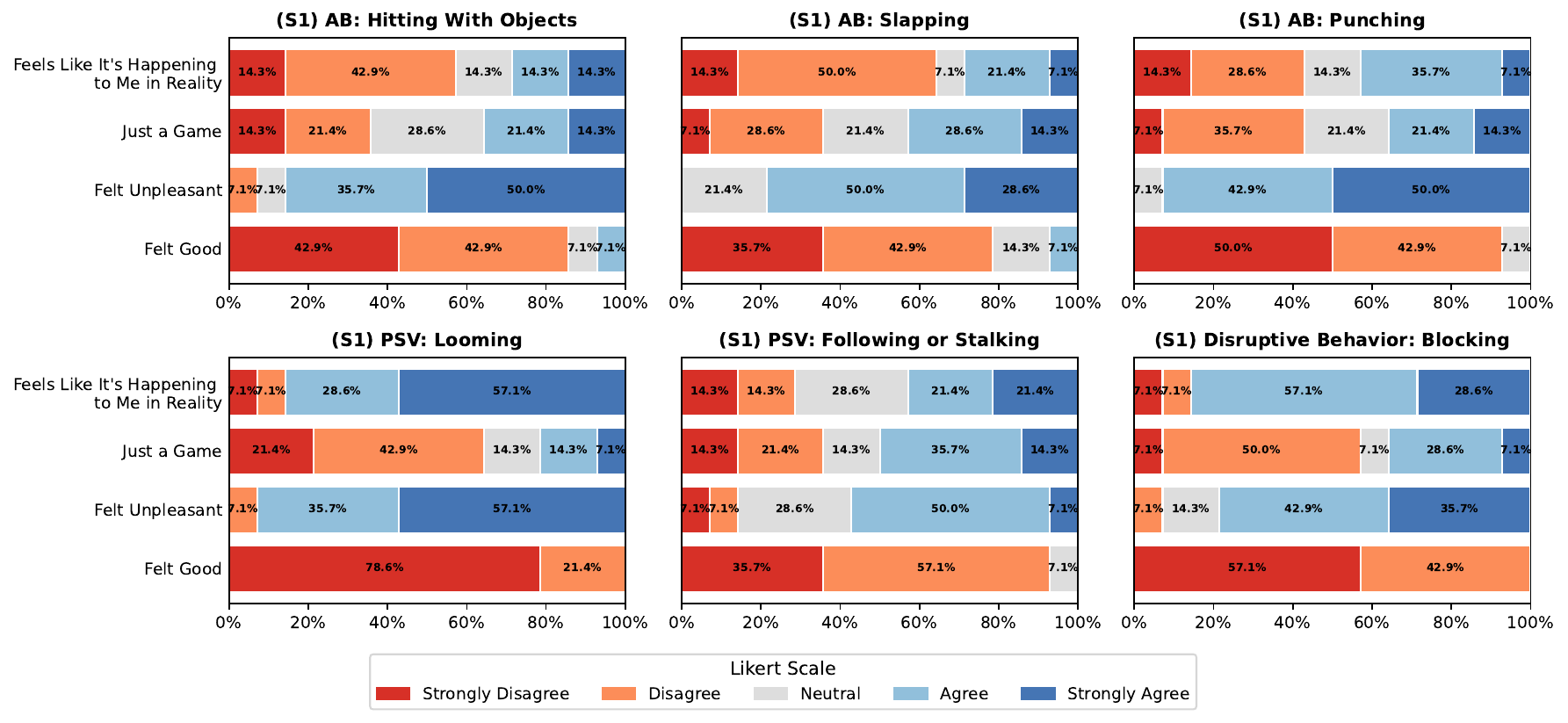}
    \caption{Participants' Likert scale responses from the Scenario 1 (S1): Communication Room (AB: Aggressive Behavior, PSV: Personal Space Violation).}
    \label{fig:communication_room}
    \vspace{-0.1in}
\end{figure*}

\begin{figure*}[!t]
    \centering
    \includegraphics[width=\linewidth]{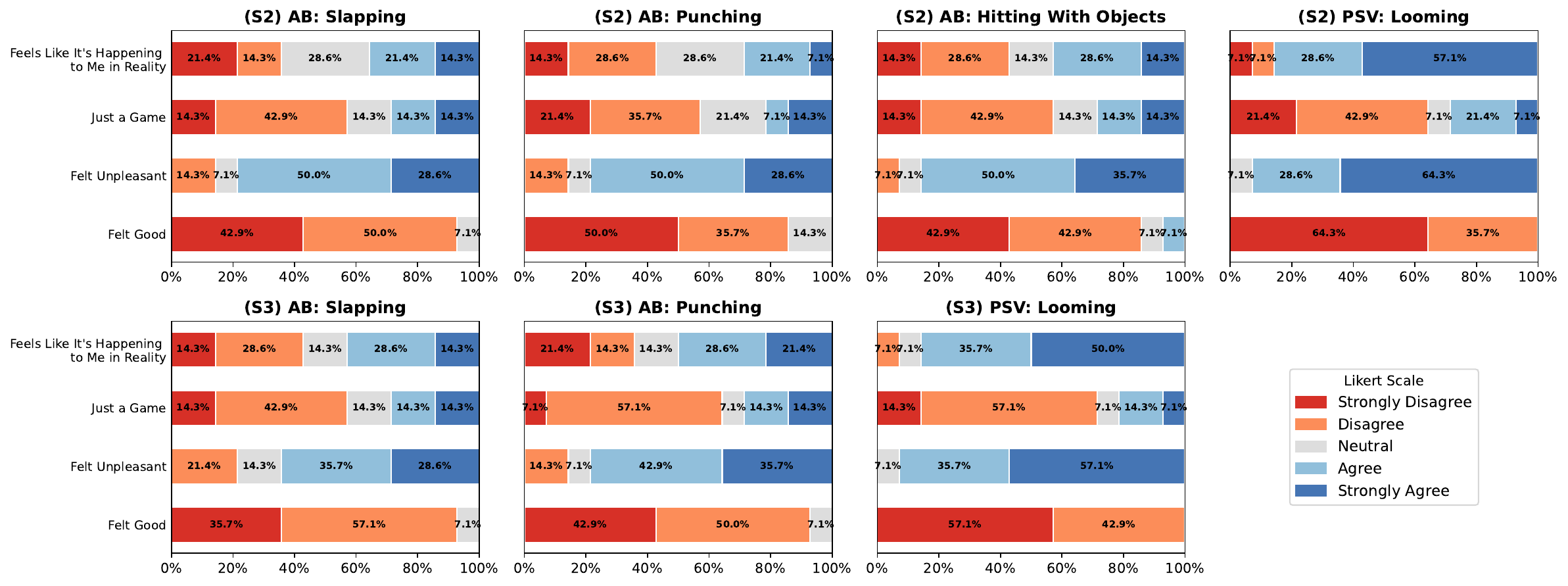}
    \caption{Participants' Likert scale responses from the Scenario 2 (S2): Whack-a-pig Room and Scenario 3 (S3): Sling Shot Room (AB: Aggressive Behavior, PSV: Personal
Space Violation).}
    \label{fig:wahckapig_slingshot_room}
    \vspace{-0.2in}
\end{figure*}

A total of 14 participants were recruited, consisting of 9 males and 5 females. Their ages ranged from 18 to 34 years. Among them, 11 participants had prior VR experience, while 3 were inexperienced. Of the experienced users (Fig.~\ref{fig:participant}), 7 reported less than one hour of VR use per day, classifying them as beginners; 3 used VR for 1–2 hours per day, corresponding to intermediate experience, and 1 used it for 2–3 hours, corresponding to advanced experience.

When asked which social VR platform they were most familiar with (i.e., VRChat, Rec Room, Meta Horizon Worlds, etc.), 12 participants answered VRChat, while 2 reported having no familiarity with any platform. However, only 4 participants had actually used a social VR platform. In contrast, all 14 participants reported experience with popular social networking apps such as Facebook, X, and Instagram. Six participants reported having witnessed harassment in either social VR platforms or other social apps. When asked whether they believed platforms take appropriate measures to address harassment, 7 participants responded ``No,'' and 6 responded ``I don’t know.'' In follow-up questions asking for reasons, participants stated:
\begin{quote}
    ``\emph{I have occasionally seen news that harassment occurs, but I have never heard what actions the platform takes.}''
    
    ``\emph{From a user perspective, I have never heard what processes are used to address incidents or whether appropriate actions were taken.}''
\end{quote}
Most participants responded that social platforms either fail to take action or that they were unaware of any actions being taken, citing a lack of transparency in how harassment cases are handled and in the outcomes of those processes.

\subsection{Results}
In this section, we analyze the survey results obtained from 14 participants in our user study.

\noindent\textbf{Differences in harassment experience by scenario.}
To examine whether the harassment experience varied by the room's context, we conducted an Analysis of Variance (ANOVA). We tested if the mean score differences between the rooms were statistically significant for participant responses to \textit{``Felt good,''} \textit{``Felt unpleasant,''} \textit{``Just a game,''} and \textit{``Feels like it's happening to me.''}

The analysis revealed that the differences between the rooms were not statistically significant for any of the response categories: ``\textit{Felt good}'' ($p = .92$, $\eta^2 = .004$), ``\textit{Felt unpleasant''} ($p = .97$, $\eta^2 = .002$), \textit{``Just a Game''} ($p = .72$, $\eta^2 = .011$), and \textit{``Feels like it's happening to me''} ($p = .81$, $\eta^2 = .016$). Specifically for the \textit{``Felt unpleasant''} response, the mean scores were highly similar across the Communication Room ($M = 4.12$, $SD=0.60$), Whack-a-Pig Room ($M = 4.14$, $SD=0.78$), and the Sling Shot Room ($M = 4.07$, $SD=0.84$).

This suggests that the user's negative experience regarding the harassment acts defined in this study is neither mitigated nor altered by the room's context. In other words, the inherent negativity of a harassing act persists across different contexts, regardless of game rules or objectives. This indicates the experience is less influenced by the scenario and more by the nature of the act itself.

\noindent\textbf{Relationship between user expertise and harassment perception.}
We performed a correlation analysis to investigate the influence of users' personal characteristics on their perception of harassment. According to the analysis, a moderate positive trend was observed between the participants' VR usage time and their realism scores for harassing acts. Although these correlations did not reach statistical significance likely due to the limited sample size, the direction of the relationship was consistent across all three room scenarios, showing similar correlation coefficients for the Communication Room ($r=.50$, $p=.09$), the Whack-a-Pig Room ($r=.46$, $p=.13$), and the Sling Shot Room ($r=.44$, $p=.15$).

This observational pattern was also evident in the responses from different participant groups. For instance, the group of participants who reported using VR for one hour or more ($N=4$) generally reported higher levels of perceived realism. Conversely, within the group of participants who had prior VR experience but used it for less than one hour ($N=8$), instances of low realism scores in the 2-point range were found across all rooms. This tendency of longer usage groups reporting higher realism and shorter usage groups reporting lower realism aligns with the correlation coefficients described above.

These findings can be interpreted as a result of VR experience enhancing users' sense of presence and immersion. Higher proficiency with the VR environment may lead users to more strongly identify with their avatars and the virtual world. This, in turn, causes external stimuli, especially harassing acts involving physical interaction, to be perceived as more realistic and direct violations. While further validation with a larger sample is needed, this analysis suggests that the experience of harassment in social VR is not solely determined by the given context but may also be moderated by the user's individual experiential characteristics.

\noindent\textbf{User perceptions across harassment categories.} Fig.~\ref{fig:communication_room} and~\ref{fig:wahckapig_slingshot_room} show participants’ Likert-scale responses. To statistically verify the differences in perception across harassment types, we conducted a repeated measures ANOVA on the four categorized behaviors (Aggressive, Stalking, Blocking, Looming). The results confirmed significant differences in perceived realism ($F(3, 39) = 8.34, p < .001, \eta_p^2 = .39$). In increasing order of perceived realism, the mean scores were as follows: Aggressive Behavior ($M = 2.88$),  Stalking ($M = 3.52$), Blocking ($M = 3.93$), and Looming ($M = 4.24$). These results indicate that participants experienced higher levels of realism when their movement or vision was constrained, compared to when they were subjected to direct strikes.

These numerical tendencies were consistently reflected in participants’ qualitative responses. One participant noted, 
\begin{quote}
``\emph{In the case of Aggressive Behavior, it was unpleasant, but felt closer to being annoying.}'' 
\end{quote}
In contrast, reactions to Looming included remarks such as, 
\begin{quote}
``\emph{Looming felt like something was really approaching me in front of my eyes, which was extremely unpleasant,}''
\end{quote}
 \begin{quote}
 ``\emph{Most of my screen was blocked, and it felt very disturbing.}''  
 \end{quote}
Such responses suggest that Looming was perceived as a more immediate and intense intrusion than the other behaviors.

Aggressive Behavior nevertheless recorded a high unpleasantness score ($M = 4.08$), consistent with the significant main effect found in the ANOVA ($F(3, 39) = 7.68, p < .001, \eta_p^2 = .37$). However, responses to the question of whether the act was perceived as \emph{``just part of the game''} showed substantial variance ($M = 2.82$, $SD = 1.18$), indicating that immersion and realism for this act were comparatively low. Blocking, on the other hand, received consistently high scores in both unpleasantness ($M = 4.07$) and realism ($M = 3.93$), much like Looming, suggesting that both behaviors were interpreted as highly realistic threats.

The strong levels of unpleasantness and realism associated with Looming and Blocking provide important insights into the nature of harassment in social VR. These behaviors appear to replicate the sense of personal space violation and loss of agency that individuals experience in the physical world. The psychological pressure and threat evoked when another person approaches uncomfortably close in real life (i.e., Looming) were nearly identical when enacted by an avatar~\cite{bailenson2003interpersonal,wilcox2006personal}. Similarly, the restriction of one’s freedom of movement (i.e., Blocking) reproduced feelings of helplessness and loss of control within the virtual space~\cite{blackwell2019harassment}.

\noindent\textbf{Harassment vs. Benign Behavior.} Although we did not collect Likert-scale ratings for benign behaviors, qualitative feedback still supports this contrast. Participants consistently distinguished harassment from benign play, explicitly noting that negative experiences arose when behaviors disrupted or violated the norms of expected gameplay. For instance, in the Whack-a-Pig and Sling Shot scenarios, participants reported that harassment:
\begin{quote}
    ``\emph{It would feel bad because it interferes with game progress."}
\end{quote}
\begin{quote}
    ``\emph{It felt annoying because it was not normal play."}
\end{quote} 
These responses suggest that benign gameplay effectively served as a neutral baseline, differentiating it from the perceived unpleasantness of harassment.

\section{HarassGuard Design}

In this section, we introduce \ourtool{}, a system that detects harassment behavior using a vision-language model (VLM). We begin by motivating the need for such a system and then present its design overview, followed by detailed descriptions.

\subsection{Technical Challenges}
\label{sec:challenges}

Our goal is to detect harassment behaviors in social VR based only on vision data. However, achieving this involves several technical challenges.

\noindent\textbf{Understanding behavioral ambiguity.} Motions, behaviors, and reactions between users (avatars) in social VR differ subtly from those in the real world. For example, when a person punches another in the physical world, the victim’s face would naturally move toward the harasser’s fist. In contrast, such realistic physical responses are not represented in avatar animations. This discrepancy creates ambiguity that makes it difficult for conventional vision models to accurately identify harassment behaviors in social VR, particularly when the models are trained on real-world harassment video datasets. Constructing large-scale VR harassment datasets would be a straightforward solution; however, this approach requires substantial data and lacks generalizability across platforms when avatars and actions are heterogeneous.

\noindent\textbf{Learning contextual information.} Detecting harassment in social VR is challenging because users are often engaged in diverse activities within virtual rooms. For example, in a `Whack-a-Pig' game room, users wield hammers to hit targets as part of the gameplay. In this context, the action of swinging a weapon is benign and necessary for the game. However, if the same action is directed toward another avatar rather than a game object, it should be identified as harassment. Conventional vision models often struggle to distinguish these nuances without contextual information, potentially misclassifying normal gameplay as aggression. Acquiring this information typically requires training on a large volume of video clips, which is costly and resource-intensive.

\subsection{System Overview}

We aim to address the above challenges by leveraging a VLM, which can process multimodal inputs such as video, images, and text. A key advantage of VLMs is their ability to incorporate contextual information—such as definitions of harassment, situational details, and action descriptions—through textual input. In this way, the need for extensive training to capture contextual information can be significantly reduced.

Fig.~\ref{fig:overview} presents the architectural overview of \ourtool{}. The system first collects video recordings of avatars’ actions from a target social VR platform and pre-processes each clip by extracting sampled frames. It then constructs a prompt containing a high-level description of the objective, the potential actions depicted in the video, and the required output format. Finally, the prompt and sampled frames are fed into the fine-tuned VLM, which classifies the video clip into a label (i.e., class) and generates a concise rationale. Next, we present the design details of prompt engineering and VLM fine-tuning.

\subsection{Prompt Engineering}
\label{subsec:prompt_engineering}

A crucial aspect of adopting a VLM as a classification model for harassment detection is the design of effective instruction prompts that enable the model to interpret both the video context and the associated harassment actions. This requires careful prompt engineering to maximize the performance of the VLM.

\noindent\textbf{Task decomposition.} To this end, we decompose our classification task into two stages: \textbf{Stage 1}, a binary classification distinguishing benign from abnormal behavior, and \textbf{Stage 2}, a multi-class classification categorizing clips into benign behavior, aggressive behavior, personal space violation, or disruptive behavior. The rationale behind this decomposition is that a VLM may fail to capture subtle distinctions among harassment behavior classes (i.e., aggressive behavior, personal space violation, and disruptive behavior) when they are combined directly with benign behavior.  

Concretely, we begin by assigning the VLM the role of \textit{``You are a strict VR harassment video classifier,''} thereby tailoring it for harassment detection in social VR. We then construct a baseline prompt for Stage 1 classification as follows:

\begin{lstlisting}[breaklines=true, basicstyle=\ttfamily\small, frame=single]
You are given {N} sequential frames sampled from ONE short VR video (chronological order).
Judge ONLY avatars' actions toward other avatars (ignore background/camera).

Benign  = normal play with no hostile intent.
Anomaly = clear hostile actions such as punching, slapping, or very close following.
If you are uncertain, choose Benign.

Return ONLY a strict JSON object with EXACTLY these fields (no extra text):
{"label": "<Benign|Anomaly>", "reason": "<one short phrase about avatars/actions/intent>"}
\end{lstlisting}
The prompt begins with a description of the input frames sampled from a VR video and the overall classification objective. It then provides concise definitions of the Benign and Anomaly classes. Finally, it enforces a strict JSON output format to ensure that the VLM does not generate extraneous text. In contrast, the Stage~2 prompt extends this design and requires the VLM to classify a VR video into four predefined categories, building on the structure of the Stage~1 prompts.

\noindent\textbf{Context augmentation.} As discussed in Section~\ref{sec:challenges}, harassment behaviors in social VR differ from those in the real world. To enable a VLM to correctly interpret VR videos, it is necessary to provide suitable contextual information. We therefore augment the baseline prompt by incorporating explicit definitions of harassment and benign behavior. To avoid biasing the VLM toward abnormal decisions (or harassment classifications in Stage~2) due to the heavier emphasis on harassment-related context—which can lower recall—we introduce a disambiguation rule. This rule instructs the VLM to classify a video as abnormal (or a harassment class) only when multiple cues are observed in the sampled frames; otherwise, the model treats the video as benign. The following snippet represents the context information added to the baseline prompt used for Stage 1 classification:
\begin{lstlisting}[breaklines=true, basicstyle=\ttfamily\small, frame=single]
Anomaly =  
1) Aggressive behavior: punching, slapping, striking with objects.  
2) Personal space violation: standing uncomfortably close, persistent following, looming.  
3) Disruptive behavior: blocking, cornering, targeted interference.  
4) If multiple weak cues occur together (e.g., following + blocking), classify as Anomaly.  
\end{lstlisting}
A full prompt text is available in our supplementary material.

\noindent\textbf{Chain-of-Thought prompting.} Another technique we adopt is chain-of-thought (CoT) prompting, which encourages an LLM to perform step-by-step reasoning before producing an answer~\cite{wei2022chain}. Applied to a VLM, CoT helps the model capture subtle distinctions among harassment behaviors in social VR, thereby improving classification performance. Accordingly, we augment the context prompt with an ordered set of reasoning steps that describe fine-grained aspects of the video relevant to harassment detection. For example, in our Stage~2 CoT prompt, we include the following steps:

\begin{lstlisting}[breaklines=true, basicstyle=\ttfamily\small, frame=single]
Reasoning (internal, do NOT output):  
1) Are multiple avatars present?  
2) Is there interaction between them?  
3) Any striking or object-based attack?  
4) Any invasive closeness, following, or looming?  
5) Any blocking, cornering, or targeted interference?  
6) If cues exist, assign the most appropriate hostile class; otherwise Benign.  
\end{lstlisting}
Note that this reasoning process is not exposed to the output to avoid verbosity. A full prompt text is available in our supplementary material.

\subsection{Fine-Tuning}

Although prompt engineering can improve model performance, VLMs may still struggle to capture the ambiguity of harassment in social VR because they are primarily trained on real-world harassment datasets. As a result, a VLM may misclassify harassment behaviors in social VR as benign, owing to differences in how avatar reactions are represented compared to real-world victims. To mitigate this issue, we fine-tune the VLM on our dataset, enabling it to learn harassment patterns specific to social VR. We uniformly sample frames from a video clip recorded and annotate them with Stage 1 and Stage 2 labels.

%% file: sections/eval.tex
\section{Evaluation}

In this section, we evaluate the effectiveness of \ourtool{} to detect harassment behavior in social VR.

\subsection{Experiment Setting}

\noindent\textbf{Dataset.}
We constructed a dataset consisting of multiple video clips (\texttt{*.mp4} files) that capture users’ actions according to the scenarios described in our user study (Section~\ref{sec:user_study}). 
Although video clips were initially collected from all 14 participants, the final dataset consists of 825 video clips from 12 participants (seven males and five females) after our validation process. This reduction occurred for the following reasons. First, all video clips from two participants (both male, aged 22, with prior VR experience but less than one hour of usage) were excluded because the majority of their recordings were severely occluded by virtual objects, such as walls or tables. Second, partial occlusions were identified in the recordings of four additional participants; however, unobstructed segments could be extracted by cropping clean video clips from the original recordings. Finally, video clips containing more than two participants were excluded, as such cases caused vision models to misrecognize one or more avatars. We plan to address this limitation in future work. The resulting 825 video clips have an average duration of 42 seconds ($\min = 10.7$, $\max = 122.1$, $SD = 21.3$). To ensure consistency, each video clip was preprocessed into 10-second segments, and clips shorter than 10 seconds were discarded. After preprocessing, we obtained a total of 3,408 video clips, whose distribution is summarized in Table~\ref{tab:dataset}. 
As detailed in Table 2, the dataset exhibits a class imbalance in Stage 2, where Aggressive Behavior (41\%) is more frequent than Disruptive Behavior (8\%). This distribution stems from the spatial constraints of the experimental setup because disruptive acts like blocking were feasible primarily in the `Communication Room' scenario, whereas aggressive actions could be enacted across all environments. We retained this distribution because the Aggressive Behavior category aggregates three sub-actions while Disruptive Behavior consists of only a single action. Consequently, the data volume per specific action does not differ significantly between the two categories.

\begin{table}[!t]
\small
\centering
\caption{Dataset distribution by Stage 1 labels, Stage 2 labels, and room categories.}
\begin{tabular}{lrr}
\toprule
\textbf{Stage 1 Labels} & \textbf{Count} & \textbf{Ratio} \\
\midrule
Anomaly & 2592 & 0.7606 \\
Benign  &  816 & 0.2394 \\
\midrule
\textbf{Stage 2 Labels} & & \\
\midrule
Aggressive Behavior                 & 1408 & 0.4131 \\
Personal Space Violation &  880 & 0.2582 \\
Disruptive Behavior       &  304 & 0.0892 \\
Benign Behavior              &  816 & 0.2394 \\
\midrule
\textbf{Rooms} \\
\midrule
Communication Room (Scenario 1) & 1864 & 0.5469 \\
Whack-a-pig Room (Scenario 2) &  728 & 0.2136 \\
Slingshot Room (Scenario 3) & 648 & 0.1901 \\
Climbing Room (Scenario 4) & 168 & 0.0493 \\
\bottomrule
\label{tab:dataset}
\end{tabular}
\vspace{-0.1in}
\end{table}

\begin{table*}[!ht]
\centering
\small
\caption{Performance of binary classification (Baselines vs. \ourtool{} under different prompting and fine-tuning settings).}
\label{tab:stage1_results}
\begin{tabular}{lcccc}
\toprule
\textbf{Model / Setting} & \textbf{Accuracy} & \textbf{Precision} & \textbf{Recall} & \textbf{F1-Score} \\
\midrule
LSTM/CNN-based (Baseline)       & 0.7704 & 0.6354 & 0.5013 & 0.4389 \\
Transformer-based (Baseline)    & 0.8804 & 0.8518 & 0.8343 & 0.8424 \\ \midrule
\ourtool{} (\textit{GPT-4o})  &        &        &        &        \\ \midrule
Baseline Prompt           & 0.6677  & 0.6888 & 0.7576 & 0.6505 \\
\quad + Context           & 0.8090       & 0.7567       & 0.8257       & 0.7730       \\
\quad + CoT               & 0.8151       & 0.7653       & 0.8394       & 0.7818       \\
\quad + Few-shot          & 0.3207       & 0.6284       & 0.5530       & 0.3034       \\ \midrule
\ourtool{} (\textit{GPT-4o, fine-tuned})   &       &        &        &        \\ \midrule
Baseline Prompt      &  0.8345      &  0.7750      & 0.8219       & 0.7919       \\
\quad + Context           & 0.8625    &  0.8094      &   0.8188     & 0.8139       \\
\quad + CoT               & 0.8809       & 0.8407       & 0.8260       & 0.8329       \\
\quad + Few-shot          &  0.6688      & 0.6898       & 0.7592       & 0.6516       \\
\bottomrule
\end{tabular}
\end{table*}

\begin{table*}[!t]
\centering
\small
\caption{Performance of multi-class classification (Baselines vs. \ourtool{} under different prompting and fine-tuning settings).}
\label{tab:stage2_results}
\begin{tabular}{lcccc}
\toprule
\textbf{Model / Setting} & \textbf{Accuracy} & \textbf{Precision} & \textbf{Recall} & \textbf{F1-Score} \\
\midrule
LSTM/CNN-based (Baseline)       & 0.4304 & 0.2327 & 0.2504 & 0.1524 \\
Transformer-based (Baseline)    & 0.7057 & 0.6758 & 0.6537 & 0.6571 \\ \midrule
\ourtool{} (\textit{GPT-4o})    &        &        &        &        \\ \midrule
Baseline Prompt    & 0.5226       & 0.5058       & 0.4817       & 0.4207       \\
\quad + Context           & 0.4827       & 0.6431       & 0.4633       & 0.3800       \\
\quad + CoT               & 0.5015       & 0.5423       & 0.4671       & 0.4075       \\
\quad + Few-shot          & 0.5467       & 0.4889       & 0.4976       & 0.4433       \\ \midrule
\ourtool{} (\textit{GPT-4o, fine-tuned})       &        &        &        &        \\ \midrule
Baseline Prompt       &  0.6574      & 0.5613       & 0.5434       & 0.5361       \\
\quad + Context           & 0.6540       & 0.5923       &  0.5578      & 0.5380       \\
\quad + CoT               & 0.6422       & 0.6205       & 0.5604       & 0.5464       \\
\quad + Few-shot          &  0.6885      & 0.6035       & 0.5716       & 0.5678       \\
\bottomrule
\end{tabular}
\vspace{-0.1in}
\end{table*}

To evaluate the effectiveness of \ourtool{}, we compare it against two conventional approaches:  
(i) a CNN (Convolutional Neural Network)/LSTM (Long Short-Term Memory) hybrid model and  
(ii) a transformer-based video classification model that does not support multimodal inputs. We evaluate performance using accuracy, precision, recall, and F1-score. All reported metrics are macro-averaged.

\noindent\textbf{CNN/LSTM-based model.}  
We implemented a model consisting of two 1D CNN layers followed by an LSTM.  
The CNN layers capture short-term repetitive behavioral patterns and extract higher-level posture features from fine-grained joint variations, with stacking increasing the receptive field.  
Compared to 2D or 3D CNNs, 1D CNNs are two to three times more efficient in terms of training and inference cost, making them preferable for this task.  
The subsequent LSTM effectively models temporal dependencies with low latency and small memory overhead, rendering the model suitable for online inference.   

\noindent\textbf{Transformer-based model.}  
We further employed a Transformer-based model designed for video classification.  
Specifically, we used the implementation provided by the Transformer library, which builds on VideoMAE—an autoencoder-based video classification framework~\cite{transfomer_model}.  
For training, we fine-tuned the model by uniformly sampling frames along the temporal axis from our dataset.  

Both the CNN/LSTM and Transformer-based models were trained and evaluated on a machine equipped with an Intel(R) Core(TM) i5-14600KF CPU, an NVIDIA GeForce RTX 4070 GPU, and 32 GB of memory.
For the CNN/LSTM model, hyperparameters were optimized using evolutionary computation, and the final configuration was selected based on the highest achieved accuracy (i.e., 128 convolution filters, 64 LSTM hidden units, 0.4 dropout rate, unidirectional LSTM, Adam optimizer, and seed 63065).
The CNN/LSTM model was trained for 100 epochs, whereas the transformer-based model was trained for 10 epochs.
In total, 1,115 video clips were used to train both models.

\noindent\textbf{Vision-language model.}
We employed GPT-4o as the VLM for \ourtool{}. The \texttt{gpt-4o-2024-08-06} model supports both image input processing and fine-tuning, making it suitable for our requirements. GPT-4o accepts images either as Base64-encoded data or through public URLs~\cite{openai_image_analysis}. Since the Base64 option incurs higher costs and reduced accuracy due to the large number of input tokens, we adopt the URL-based approach. Specifically, we upload preprocessed frames (sampled from video clips) to a Cloudflare R2 bucket and embed the resulting public URLs into our instruction prompts (Section~\ref{subsec:prompt_engineering}). This setup allows GPT-4o to retrieve and analyze six sampled frames per video clip. All uploaded frames are anonymized and contain no privacy-sensitive information.

\noindent\textbf{VLM fine-tuning.} For fine-tuning, we randomly sampled 200 videos from the dataset while preserving the label distribution. We fine-tuned \texttt{GPT-4o-2024-08-06} on approximately 2.37M task-specific tokens using supervised fine-tuning for 3 epochs, resulting in a stable model used in our evaluation.

\noindent\textbf{Comparsion of model latency.} We further measured the average inference time per 10-second video clip. The LSTM/CNN baseline demonstrated the highest efficiency with an average inference time of 1.40 seconds, followed by the Transformer-based model at 3.16 seconds. In contrast, \ourtool{} recorded an average of 5.52 seconds due to the processing overhead of the large-scale VLM.

\subsection{Performance of Binary Classification (Stage 1)}

Table~\ref{tab:stage1_results} presents the performance of binary classification (Stage 1) across baseline models and \ourtool{} (\textit{GPT-4o}). For the baseline models, the Transformer-based classifier clearly outperformed the LSTM/CNN-based model, achieving higher accuracy (88.04\% vs. 77.04\%) and a substantially better F1-score (0.8424 vs. 0.4389). This indicates that Transformer architectures better capture the temporal and contextual patterns in video frames relevant to harassment behaviors.

We next evaluate \ourtool{} under different prompting and fine-tuning settings. Prior to fine-tuning, \ourtool{} exhibited modest performance when using the baseline prompt alone (66.77\% accuracy, F1-score 0.6505). Incorporating contextual information into the prompt substantially improved performance, increasing accuracy to 80.90\% and F1-score to 0.7730, with recall improving from 75.76\% to 82.57\%. This improvement highlights the importance of explicitly providing contextual cues when applying multimodal language models to behavioral classification tasks.

In contrast, few-shot prompting led to a severe performance drop (32.07\% accuracy), indicating that naive example-based prompting is insufficient for this task. Adding Chain-of-Thought (CoT) reasoning proved more effective, achieving 81.51\% accuracy and an F1-score of 0.7818, which surpassed the performance of the contextual prompt. This suggests that guiding the model through a step-by-step reasoning process is effective for this classification task.

After fine-tuning, \ourtool{} consistently improved across all prompt settings. The baseline prompt reached 83.45\% accuracy (F1-score 0.7919), while contextual prompting further improved performance to 86.25\% accuracy (F1-score 0.8139). The strongest performance was achieved by combining CoT prompting with fine-tuning (88.09\% accuracy, F1-score 0.8329), closely matching the transformer-based baseline. Few-shot prompting remained ineffective even after fine-tuning, reinforcing that context-rich and reasoning-enhanced prompts are more suitable than naive few-shot examples for this task.

\subsection{Performance of Multi-class Classification (Stage 2)}
Table~\ref{tab:stage2_results} shows the performance of multi-class classification (Stage 2), where the task is to categorize video frames into fine-grained harassment behaviors, such as Aggressive Behavior, Personal Space Violation, Disruptive Behavior, and Benign Behavior.

Compared to Stage 1, overall performance decreased considerably across all models, reflecting the increased difficulty of distinguishing multiple specific behavior classes. The LSTM/CNN-based baseline achieved only 43.04\% accuracy and an F1-score of 0.1524, while the Transformer-based model performed better (70.57\% accuracy, F1-score 0.6571) but still showed room for improvement. This indicates that fine-grained behavioral classification is challenging even for strong sequence models.

For \ourtool{} (\textit{GPT-4o}) without fine-tuning, baseline prompting achieved moderate performance (52.26\% accuracy, F1-score 0.4207). Adding context or CoT reasoning did not consistently improve results; for example, the context prompt slightly decreased accuracy to 48.27\%, and CoT achieved 50.15\%. Few-shot prompting yielded the highest accuracy among the zero-shot variants (54.67\%), suggesting that providing a few labeled examples may help the model partially disambiguate fine-grained classes, although the F1-score remains modest (0.4433).

After fine-tuning, \ourtool{} showed noticeable improvements, with baseline prompt accuracy reaching 65.74\% (F1-score 0.5361). Few-shot prompting after fine-tuning achieved the best performance overall (68.85\% accuracy, F1-score 0.5678), indicating that combining fine-tuning with minimal examples allows the model to better capture distinctions among detailed behaviors. Interestingly, CoT prompting after fine-tuning did not outperform the baseline or few-shot prompts, suggesting that reasoning-style prompts are less effective than direct exposure to labeled examples for this task.

\subsection{Effect of Prompting vs. Fine-tuning}
Overall, these findings indicate that prompt engineering, particularly adding context, substantially improved Stage 1 performance. Accuracy increased from 66.77\% with the baseline prompt to 80.90\% when using the context prompt, and F1-score rose from 0.6505 to 0.7730. CoT reasoning provided additional gains, achieving 81.51\% accuracy and an F1-score of 0.7818. In particular, applying these prompts to the zero-shot model allowed it to approach the performance of fine-tuned variants. The fine-tuned variants improved results to 88.09\% accuracy and an F1-score of 0.8329 when combining with CoT, closely matching the Transformer baseline.

In Stage 2, unlike Stage 1, adding context or CoT prompts without fine-tuning did not consistently improve performance over the baseline prompt. Context prompting slightly decreased accuracy to 48.27\% (F1-score 0.3800), while CoT prompting achieved 50.15\% accuracy (F1-score 0.4075). Few-shot prompting provided a modest increase, reaching 54.67\% accuracy and an F1-score of 0.4433. Fine-tuning substantially increased performance, with the baseline prompt reaching 65.74\% accuracy and an F1-score of 0.5361, and combining fine-tuning with few-shot examples resulting in 68.85\% accuracy and an F1-score of 0.5678.

%% file: sections/relatedwork.tex
\section{Discussion and Limitations}

\noindent\textbf{Ethical Consideration}. Prior to the experiment, participants were fully informed about the nature of the tasks and the potential for psychological discomfort. We explicitly notified them that they could voluntarily withdraw from the study at any time without any penalty. In the event of early withdrawal, participants were still compensated with 50\% of the total reward for their time. To mitigate fatigue, mandatory five-minute breaks were provided between scenarios, with additional breaks granted immediately upon participants' request.

\noindent\textbf{Low performance in multi-class classification.}
Although we decomposed the classification task into two stages to improve accuracy, our results show that \ourtool{} underperformed on multi-class classification compared with the Transformer-based model. We acknowledge this limitation, but we anticipate that fine-tuning the VLM on a larger dataset could substantially improve its performance. For instance, GPT-4o was fine-tuned with only 200 video samples due to budget constraints, whereas the Transformer-based model was trained with about 1,000 samples. We conjecture that this imbalance contributed to the VLM’s weaker Stage 2 performance. Nevertheless, given the limited number of training samples, we argue that the VLM still achieved notable improvements (31.85\% over the baseline without fine-tuning), demonstrating its potential for further scaling.

\noindent\textbf{Lack of avatar diversity.} In this study, we focused on a single humanoid avatar as a proof-of-concept to validate the feasibility of VLM-based harassment detection in social VR. While this choice allowed us to control visual variability and isolate behavioral cues, it limits generalizability to environments with heterogeneous avatar designs. In parallel experiments with LSTM/CNN models, we explored pose estimation tools (e.g., MediaPipe~\cite{mediapipe}, YOLOv8~\cite{yolov8}) to convert visual inputs into skeletal data to mitigate avatar dependency. Building on these insights, future work will evaluate \ourtool{} across a broader range of avatar styles (e.g., non-humanoid, stylized, or animal-like avatars) while implementing skeleton-based preprocessing to standardize heterogeneous inputs, aiming to assess robustness under realistic deployment conditions.

\noindent\textbf{Local models and latency optimization.} The primary objective of this work is to validate the feasibility of VLM-based harassment detection in social VR as a proof-of-concept, rather than to conduct an exhaustive comparison of different VLM architectures (e.g., LLaMA, DeepSeek). Accordingly, we selected GPT-4o to establish a strong performance baseline, leveraging its advanced capabilities for multimodal reasoning and contextual understanding. Although our cloud-based implementation introduces network latency that may limit real-time applicability, this does not undermine our core contribution—demonstrating that VLMs can effectively reason about complex harassment behaviors using visual input alone. Future work will address this by evaluating \ourtool{} with locally deployable VLMs to optimize inference latency and throughput for practical deployment.

%% file: sections/conclusion.tex
\section{Conclusion}

Online harassment in social VR has recently emerged as a serious problem, yet no dedicated dataset exists, and existing solutions are not tailored for harassment detection in this domain due to the lack of contextual information. In this paper, we present a user study to construct a vision dataset grounded in a systematic classification of harassment behaviors in social VR. Building on this dataset, we propose \ourtool{}, a novel detection system that leverages a vision-language model (VLM) to classify harassment behaviors with richer context through multimodal video and text inputs. Our experiments show that \ourtool{} effectively detects harassment behaviors when the VLM is enhanced with prompt engineering and fine-tuning.

\acknowledgments{
This work was partially supported by Institute of Information \& communications Technology Planning \& Evaluation (IITP) [RS-2023-00215700, Trustworthy Metaverse: blockchain-enabled convergence research, 70\%] and the National Research Foundation of Korea (NRF) grant funded by the Korea government (MSIT) [RS-2024-00457937, 30\%].
}